%% file: main.tex
\documentclass[journal]{IEEEtran}
\usepackage[T1]{fontenc}
\usepackage{tgtermes}
\usepackage{amsmath,amssymb}
\usepackage{graphicx}
\usepackage{booktabs}
\usepackage{array}
\usepackage{xcolor}
\usepackage{microtype}
\usepackage{cite}
\usepackage{algorithm}
\usepackage{algorithmic}
\usepackage[hidelinks]{hyperref}
\input{numbers}
\input{numbers2}

\newcommand{\cam}{\mathcal{L}_{\mathrm{cam}}}
\newcommand{\dt}{\mathcal{L}_{\mathrm{dt}}}
\newcommand{\lr}{\ell_{\mathrm{LR}}}
\newcommand{\fh}{\hat{\mathbf{f}}}

\hypersetup{pdftitle={Separating Protocol Effects from Spatio-Temporal Evidence in Person Re-Identification Re-Ranking},
  pdfauthor={Mochamad Rizal Fauzan, Doan Pham Thao Nhu, Jeslin Konthoujam, Purit Hongjirakul, Lakshay Chawla},
  pdfkeywords={person re-identification; spatio-temporal prior; re-ranking; evaluation protocol; likelihood ratio; reproducibility}}

\begin{document}

\title{Separating Protocol Effects from Spatio-Temporal\\
Evidence in Person Re-Identification Re-Ranking}

\newcommand{\authorbox}[2]{\begin{tabular}[t]{@{}c@{}}
#1\\[2pt]
{\footnotesize National Taipei University of Technology}\\
{\footnotesize Taipei, Taiwan}\\
{\footnotesize #2}
\end{tabular}}
\author{%
\authorbox{Mochamad Rizal Fauzan}{t114998415@ntut.edu.tw}\hspace{0.9cm}%
\authorbox{Doan Pham Thao Nhu}{t114998401@ntut.edu.tw}\hspace{0.9cm}%
\authorbox{Jeslin Konthoujam}{t114998405@ntut.edu.tw}\\[9pt]
\authorbox{Purit Hongjirakul}{t114998411@ntut.edu.tw}\hspace{0.9cm}%
\authorbox{Lakshay Chawla}{t113998405@ntut.edu.tw}}

\markboth{Fauzan \MakeLowercase{\textit{et al.}}: Separating Protocol Effects from Spatio-Temporal Evidence in Person Re-Identification Re-Ranking}{Fauzan \MakeLowercase{\textit{et al.}}: Separating Protocol Effects from Spatio-Temporal Evidence in Person Re-Identification Re-Ranking}

\maketitle

\begin{abstract}
Spatio-temporal priors re-rank person re-identification matches using the camera pair and the time gap, and are reported to add large gains on Market-1501.
We ask how much of such a gain encodes the evaluation protocol rather than learned spatio-temporal evidence.
Starting from a course-project re-ranker that adds log camera-transition and log time-gap terms to cosine similarity, we train a strong ResNet-50 baseline (Rank-1 \BaseRone\,\%, mAP \BaseMap\,\% over three seeds), choose the fusion weights on an identity-disjoint validation split and evaluate on the official test set with control priors.
The repository form adds \GorigMapM\ mAP points, but a one-parameter penalty on same-camera pairs alone adds \GscMapM: the protocol never places a true match in the query's camera, so the penalty removes distractors for free.
The time term of that form is almost unused; a pair-level diagnostic traces this to its modelling the time gap given a match without dividing by the gap distribution of all candidate pairs.
A likelihood-ratio prior adds \GlrMapM\ points without the penalty and \GlrpMapM\ with it; shuffling the time stamps removes the extra gain, so about \TimeBeyondProtocolM\ points are tied to identity.
The prior fixes the first-ranked match of \QualPctimproved\,\% of the queries and breaks it for \QualPctworsened\,\%.
The gain shrinks as the visual baseline improves (\EpTwentylrp\ to \GlrpMapM\ points between a \EpTwentyBase\,\% and a \BaseMap\,\% mAP model) and concentrates on queries whose true matches share the query's video sequence.
The study covers one dataset, one backbone and three seeds.
\end{abstract}


\input{sections/intro}
\input{sections/related}
\input{sections/method}
\input{sections/setup}
\input{sections/results}
\input{sections/discussion}
\input{sections/conclusion}
\input{sections/appendix}

\section*{Data and Code Availability}
Market-1501 \cite{zheng2015scalable} is not redistributed here and must be obtained from its authors.
The code that trains the baselines, selects the fusion weights, runs every control and generates every number, table and figure of this paper is in the repository \url{https://github.com/rizalfanex/spatiotemporal-bayesian-reid}, folder \texttt{paper\_audit}.
Everything reproduces from the training commands (a validation-mode and a full-mode run per seed, about 15 minutes each on one consumer GPU, about 1.5 hours for three seeds), one analysis script that selects the weights and writes the result tables, and the scripts that build the numbers, figures and references of this paper from those files; the per-query average precisions of every configuration are stored so that the bootstrap intervals can be recomputed.

\input{references}
\end{document}

%% file: numbers.tex
\newcommand{\BaseMapPm}{86.18\,$\pm$\,0.24}
\newcommand{\BaseRonePm}{94.27\,$\pm$\,0.05}
\newcommand{\BaseMap}{86.2}
\newcommand{\BaseRone}{94.3}

\newcommand{\GorigMapM}{2.30}
\newcommand{\GorigOneM}{2.21}

\newcommand{\GscMapM}{2.19}

\newcommand{\GorigshMapM}{2.13}

\newcommand{\GlrMapM}{4.04}

\newcommand{\GlrpMapM}{5.62}
\newcommand{\GlrpOneM}{4.11}

\newcommand{\GlrshMapM}{0.00}

\newcommand{\GlrpshMapM}{2.19}

\newcommand{\AbsLrpOne}{98.4}

\newcommand{\TimeBeyondProtocolM}{3.43}
\newcommand{\ProtocolShare}{39}
\newcommand{\Ciorig}{[2.15, 2.46]}

\newcommand{\Cisc}{[2.04, 2.33]}

\newcommand{\Cilr}{[3.69, 4.39]}

\newcommand{\Cilrp}{[5.21, 6.02]}

\newcommand{\CilrpMinusOrig}{[3.00, 3.65]}
\newcommand{\BmlrpMinusOrig}{3.31}
\newcommand{\CilrpMinusSc}{[3.10, 3.77]}

\newcommand{\CilrMinusSh}{[3.70, 4.40]}
\newcommand{\BmlrMinusSh}{4.04}
\newcommand{\NQuery}{3{,}368}
\newcommand{\NBoot}{5000}
\newcommand{\StrShareall}{18.6}

\newcommand{\StrBaseall}{77.5}
\newcommand{\Strallorig}{+3.0}

\newcommand{\Strallsc}{+3.1}

\newcommand{\Stralllr}{+13.0}

\newcommand{\Stralllrp}{+14.6}

\newcommand{\StrSharesome}{67.0}

\newcommand{\StrBasesome}{88.3}
\newcommand{\Strsomeorig}{+2.1}

\newcommand{\Strsomesc}{+1.9}

\newcommand{\Strsomelr}{+2.5}

\newcommand{\Strsomelrp}{+3.9}

\newcommand{\StrSharenone}{14.4}

\newcommand{\StrBasenone}{87.4}
\newcommand{\Strnoneorig}{+2.2}

\newcommand{\Strnonesc}{+2.2}

\newcommand{\Strnonelr}{-0.5}
\newcommand{\StrnonelrCi}{[-0.8, -0.2]}
\newcommand{\Strnonelrp}{+1.9}

\newcommand{\NValQ}{413}
\newcommand{\NValG}{1{,}181}

\newcommand{\AucRaw}{0.720}
\newcommand{\AucSeq}{0.663}
\newcommand{\AucCam}{0.586}
\newcommand{\AucDelta}{0.560}
\newcommand{\AucBoth}{0.601}
\newcommand{\AucLR}{0.841}
\newcommand{\DiagPairs}{1{,}024{,}728}
\newcommand{\DiagPos}{12{,}641}
\newcommand{\DiagPosPct}{1.23}
\newcommand{\ShareSeqPos}{59}
\newcommand{\ShareSeqNeg}{26}
\newcommand{\AucWithinRaw}{0.860}
\newcommand{\AucWithinDelta}{0.622}
\newcommand{\AucWithinLR}{0.885}
\newcommand{\EpTwentyBase}{70.5}
\newcommand{\EpTwentyorig}{5.3}

\newcommand{\EpTwentysc}{4.7}

\newcommand{\EpTwentylr}{8.4}

\newcommand{\EpZerolr}{13.7}
\newcommand{\EpTwentylrp}{11.9}

\newcommand{\EpZeroBase}{2.7}
\newcommand{\LrpBetaSel}{0.02, 0.03, 0.03}
\newcommand{\LrpGammaSel}{0.03, 0.03, 0.05}
\newcommand{\OrigGammaSel}{0.005, 0.005, 0}

%% file: numbers2.tex
\newcommand{\CmcLrpFive}{99.5}
\newcommand{\CmcBaseFive}{98.2}
\newcommand{\CmcLrpTen}{99.7}
\newcommand{\CmcBaseTen}{98.9}
\newcommand{\SensMin}{5.44}
\newcommand{\SensMax}{5.94}
\newcommand{\CamPRMin}{1.4}
\newcommand{\CamPRMax}{3.6}
\newcommand{\CamLowBaseAP}{80.2}
\newcommand{\CamHighBaseAP}{89.7}
\newcommand{\CamGainMin}{4.1}
\newcommand{\CamGainMax}{7.8}
\newcommand{\Qualimproved}{140}
\newcommand{\QualPctimproved}{4.2}
\newcommand{\Qualworsened}{4}
\newcommand{\QualPctworsened}{0.1}
\newcommand{\Qualbothcorrect}{3{,}170}

\newcommand{\Qualbothwrong}{54}
\newcommand{\QualPctbothwrong}{1.6}

\newcommand{\LandValMax}{2.08}
\newcommand{\LandTestAtSel}{5.80}
\newcommand{\LandTestMax}{5.80}
\newcommand{\LandBeta}{0.02}
\newcommand{\LandGamma}{0.03}
\newcommand{\NGalleryTest}{15{,}913}
\newcommand{\TimeSim}{6.7}
\newcommand{\TimeTerms}{13.5}
\newcommand{\TimeFuse}{2.6}
\newcommand{\TimeEval}{17.8}
\newcommand{\TimeTable}{58.4}
\newcommand{\TimeGpu}{RTX 5080}
\newcommand{\TimeFusionTotal}{16.1}
\newcommand{\NPosShort}{114{,}025}
\newcommand{\NAllShort}{68{,}090{,}304}

%% file: sections/intro.tex
\section{Introduction}
\IEEEPARstart{P}{erson} re-identification (re-ID) retrieves, for a query image of a pedestrian seen by one camera, the images of the same person captured by other cameras of a network \cite{zheng2016past,ye2020deep}.
It is a building block of multi-camera tracking \cite{ristani2018features,bewley2016simple} and of person search \cite{xiao2016joint}, and the benchmark progress of the last decade, from hand-crafted descriptors and metric learning \cite{gray2008viewpoint,farenzena2010person,zheng2011person,koestinger2012large,liao2015person} to deep embeddings \cite{li2014deepreid,zheng2015scalable,sun2017beyond,zhou2019omni} and transformers \cite{he2021transreid}, has been driven almost entirely by better visual features.

Appearance alone is ambiguous when many people look alike, but a camera network also records \emph{where} and \emph{when} each image was taken.
Such information is cheap, available at deployment, and informative: a pedestrian leaving one camera is likely to appear in a neighbouring camera after a plausible delay \cite{makris2004bridging,javed2008modeling,cho2017distance}.
Spatio-temporal priors therefore re-rank the visual matches by the probability of the camera pair and the time gap, and have been reported to raise Market-1501 \cite{zheng2015scalable} from 91.2\,\% to 98.1\,\% Rank-1 on top of a part-based baseline \cite{wang2018spatial}.

A prior of this kind is easy to get wrong in ways that still raise the benchmark number.
We started from a course-project implementation (the repository named in the availability statement) that adds a log camera-transition term and a log time-gap term to the cosine similarity.
Its baseline was weak (Rank-1 46.3\,\%, mAP 23.2\,\%) and its fusion weights had been chosen on the test set, so its reported gain could not be trusted as it stood.
The evaluation protocol of the dataset adds a second, subtler risk: gallery images of the query's identity \emph{and} camera are ignored, so no true match is ever in the query's camera, and any term that penalises same-camera pairs exploits this knowledge of the protocol without learning anything from data.
Rather than polish a method, we turned the project into an audit and asked four questions.
\begin{enumerate}
\item[\textbf{Q1}] \textbf{Protocol.} How much of the gain comes from a term that merely knows the evaluation protocol?
\item[\textbf{Q2}] \textbf{Identity link.} Is the remainder tied to the identity of the person, or would any prior that reshapes the score distribution do as well? We test this by shuffling the time stamps when the prior is built.
\item[\textbf{Q3}] \textbf{Baseline strength.} Does the gain persist when the visual model is strong?
\item[\textbf{Q4}] \textbf{Where.} For which queries, and which cameras, does the gain appear?
\end{enumerate}

\noindent The contributions of this paper are as follows.
\begin{itemize}
\item A decomposition of the gain of an additive camera and time re-ranker into a same-camera penalty and a learned part, with a time-shuffled control for each prior and a validation protocol in which no weight is tuned on the test set. The penalty alone reproduces \GscMapM\ of the \GorigMapM\ mAP points of the repository form.
\item A pair-level diagnostic (AUC \AucDelta\ for the repository's time term against \AucLR\ for a likelihood-ratio prior) that explains why that term carries little signal: it models the time gap given a match but is not normalised by the gap distribution of all candidate pairs.
\item An evaluation of the likelihood-ratio prior, closely related to the conditional match probability of Wang \emph{et al.}\ \cite{wang2018spatial}, which adds \GlrMapM\ mAP points without and \GlrpMapM\ with the penalty on a strong baseline; about \TimeBeyondProtocolM\ points beyond the penalty vanish under time shuffling.
\item A study of how the gain depends on baseline strength, on the temporal closeness of the true matches, on the query camera and on the design of the prior, with quantitative and qualitative evidence, a full system description and open code.
\end{itemize}
We do not propose a new state-of-the-art method, and we did not re-implement the full model of \cite{wang2018spatial}; our claims concern the additive log-prior fusion that we studied.

The remainder of the paper is organised as follows.
Section~\ref{sec:related} reviews related work, Section~\ref{sec:method} formulates the problem and describes the audited system, Section~\ref{sec:setup} gives the experimental setup, Section~\ref{sec:results} reports results, and Sections~\ref{sec:discussion} and~\ref{sec:conclusion} discuss limitations and conclude.

%% file: sections/related.tex
\section{Related Work}
\label{sec:related}
\subsection{Person Re-Identification and Its Benchmarks}
Early re-ID relied on hand-crafted appearance descriptors with learned metrics \cite{gray2008viewpoint,farenzena2010person,zheng2011person,koestinger2012large,liao2015person}.
Deep filter-pairing networks \cite{li2014deepreid} and the large Market-1501 benchmark \cite{zheng2015scalable} moved the field to learned embeddings; later benchmarks and generative tools such as Person Transfer GAN \cite{wei2018person}, which also provides the MSMT17 dataset, increased scale and diversity.
Architectural work has explored part-based pooling \cite{sun2017beyond,fu2018horizontal,zheng2018pyramidal}, multiple granularities \cite{wang2018learning}, attention \cite{li2018harmonious,chen2019mixed}, multi-level factorisation \cite{chang2018multi}, omni-scale features \cite{zhou2019omni,zhou2019learning}, architecture search \cite{quan2019auto}, and, more recently, vision transformers \cite{dosovitskiy2020image,he2021transreid}, self-supervised pre-training \cite{luo2021self} and vision--language models \cite{li2022clip}.
Surveys \cite{zheng2016past,ye2020deep} and the study of re-ID under realistic detection conditions \cite{zheng2016wild} summarise the landscape.
All of these improve the visual score; none changes the observation that appearance is ambiguous across cameras, which motivates the priors we audit.

\subsection{Training Recipes and Strong Baselines}
A strong baseline is a precondition for judging any add-on.
Luo \emph{et al.}\ \cite{luo2019bag} collect training tricks for a global-feature ResNet-50 \cite{he2015deep} and report 94.5\,\% Rank-1 and 85.9\,\% mAP on Market-1501.
The ingredients are common in the wider literature: metric-learning losses with hard-negative mining \cite{schroff2015facenet,hermans2017defense,wu2017sampling,chen2017beyond}, label smoothing \cite{szegedy2015rethinking,muller2019does}, batch normalisation \cite{ioffe2015batch}, ImageNet pre-training \cite{russakovsky2014imagenet}, the Adam optimiser \cite{kingma2015adam} and its decoupled-decay variant \cite{loshchilov2017decoupled}, data augmentation by random erasing and cutout \cite{zhong2017random,devries2017improved}, and mixed-precision training \cite{micikevicius2017mixed,kalamkar2019study}.
Our baseline follows this family and is implemented in PyTorch \cite{paszke2019pytorch}; Section~\ref{sec:setup} lists the settings, which we did not copy from \cite{luo2019bag}.

\subsection{Domain Adaptation and Unsupervised Re-ID}
Camera- and time-dependent statistics are properties of a deployment, which links priors to adaptation.
Cross-dataset transfer \cite{peng2016unsupervised}, image-to-image translation that preserves identity \cite{deng2018image,zhong2018camera}, generated samples \cite{zheng2017unlabeled,zheng2019joint}, clustering and pseudo-label refinement \cite{fan2017unsupervised,ge2020mutual,ge2020self,dai2021cluster} and exemplar memories \cite{zhong2019invariance} address the visual side of the shift between camera networks.
A spatio-temporal prior estimated on one network does not transfer automatically to another, a point we return to in Section~\ref{sec:discussion}.

\subsection{Spatio-Temporal and Camera-Topology Priors}
Space--time relations between non-overlapping cameras were modelled for tracking before deep re-ID \cite{makris2004bridging,javed2008modeling}, and recent work infers camera network topology from distances \cite{cho2017distance}.
Wang \emph{et al.}\ \cite{wang2018spatial} estimate, from training pairs, the probability that two images show the same person given the time-gap bin and the two cameras, smooth the histograms with a Parzen window and combine the visual and spatio-temporal scores with a logistic joint metric.
Other work transfers spatial-temporal patterns across datasets \cite{lv2018unsupervised}, learns instance-level spatial-temporal patterns \cite{ren2021learning}, fuses spatial-temporal networks with causal matching \cite{kim2024object}, mines auxiliary information from datasets for unsupervised re-ID \cite{teng2022high}, or builds a large benchmark in which spatio-temporal information is central \cite{shu2021large}.
We do not re-implement these models.
Our study concerns a simpler additive fusion of log priors with cosine similarity, and asks what that kind of fusion adds on a strong baseline once the protocol is accounted for.
Re-ranking of the visual ranking list by k-reciprocal encoding \cite{zhong2017re} or by cross-neighbourhood expansion \cite{sarfraz2018pose} needs no camera or time information and addresses a different part of the problem; we do not combine it with the priors studied here.

\subsection{Benchmark Pitfalls and Reporting Practice}
The risk that a benchmark rewards cues that its protocol makes available, rather than the capability it is meant to measure, is general in machine learning: shortcut learning \cite{geirhos2020shortcut}, test-set reuse, label quality and distribution shift \cite{recht2019imagenet,beyer2020done,koh2021wilds}, the choice of benchmarks \cite{raji2021everything,dehghani2021benchmark}, and weak baselines and tuning on the test set \cite{henderson2018deep,lipton2018troubling,musgrave2020metric}.
In metric learning, a careful re-evaluation shows that claimed improvements over older methods can shrink or vanish under fair protocols \cite{musgrave2020metric}.
Our same-camera control is a concrete instance of this concern for re-ID evaluation, and our likelihood-ratio form treats the prior as a probability estimated from data and not as an arbitrary score offset; the calibration of the visual score itself is a separate question \cite{guo2017calibration} that we do not address.

%% file: sections/method.tex
\section{Problem Formulation and the Audited System}
\label{sec:method}
\subsection{Task, Protocol and Notation}
A query $q$ and a gallery image $g$ come with a camera index $c\in\{1,\dots,6\}$, a sequence number and a frame number.
Let $y_{qg}\in\{0,1\}$ indicate that the two images show the same person.
The standard Market-1501 protocol ranks all gallery images for each query, ignores gallery images that have both the identity and the camera of the query, and reports Rank-$k$ accuracy and mean average precision (mAP).
Table~\ref{tab:notation} lists the notation used throughout the paper.
\begin{table}[!t]
\caption{Notation.}
\label{tab:notation}
\centering
\footnotesize\setlength{\tabcolsep}{4pt}
\begin{tabular}{@{}ll@{}}
\toprule
Symbol & Meaning \\ \midrule
$q,\ g$ & query image and gallery image \\
$c\in\{1,\dots,6\}$ & camera index of an image \\
$t,\ \Delta t=|t_q-t_g|$ & pseudo-time of an image and gap between two images \\
$b,\ x=(c_q,c_g,b)$ & gap bin and the spatio-temporal cell of a pair \\
$y_{qg}\in\{0,1\}$ & 1 if $q$ and $g$ show the same person \\
$\fh,\ s_{qg}=\fh_q^{\top}\fh_g$ & L2-normalised feature and cosine similarity \\
$\cam,\ \dt,\ \lr$ & camera term, time term and log likelihood ratio \\
$\beta,\ \gamma$ & fusion weights chosen on validation \\
$\varepsilon,\ \alpha$ & floor of the same-camera penalty and smoothing count \\
$\tilde{s}_{qg}$ & fused score used for ranking \\ \bottomrule
\end{tabular}
\end{table}
Two consequences matter for what follows.
First, every counted true match is in a camera different from the query's.
Second, gallery images of the query's camera with a different identity are \emph{not} ignored: they remain distractors that a visual score must rank below the true matches.

The pseudo-time of an image is $t=(\text{seq}-1)\cdot 200{,}000+\text{frame}$, using the numbers in the Market-1501 file names; this offset is the convention of the repository we started from and puts different sequences far apart.
The gap $\Delta t=|t_q-t_g|$ is therefore fine-grained within a sequence and coarse across sequences.

\subsection{A Bayesian View of the Fusion}
Let $v$ denote the visual evidence of a pair and $x=(c_q,c_g,b)$ its spatio-temporal evidence, with $b$ the bin of $\Delta t$.
If $v$ and $x$ are conditionally independent given $y$, Bayes' rule gives the log posterior odds
\begin{multline}
\log\frac{P(y{=}1\mid v,x)}{P(y{=}0\mid v,x)}
= \log\frac{P(v\mid y{=}1)}{P(v\mid y{=}0)}\\
+ \log\frac{P(x\mid y{=}1)}{P(x\mid y{=}0)}
+ \log\frac{P(y{=}1)}{P(y{=}0)},
\label{eq:bayes}
\end{multline}
the sum of a visual term, a spatio-temporal term and the prior odds.
The cosine similarity $s_{qg}$ is a monotone surrogate of the first term, and its scale relative to the second is unknown, so the second term enters with a weight $\gamma$ chosen on validation data.
Matches are about \DiagPosPct\,\% of the cross-camera pairs, so $P(x\mid y{=}0)\approx P(x)$ and the second term is estimated as a ratio against the distribution of \emph{all} candidate pairs.
Equation~\eqref{eq:bayes} tells what a correct time term looks like: it is a ratio, and a term that models only $P(x\mid y{=}1)$ rewards time gaps that are common among all pairs, whether or not they separate matches from non-matches.
The constant last term does not change the ranking within a query.

\subsection{Fusion and the Priors Under Audit}
The re-ranker scores each pair as
\begin{equation}
\tilde{s}_{qg} = s_{qg} + \beta\,\cam(q,g) + \gamma\,\dt(q,g),\qquad s_{qg}=\fh_q^{\top}\fh_g,
\label{eq:fuse}
\end{equation}
where $\fh$ is the L2-normalised visual feature, $\cam$ and $\dt$ are log priors built only from training identities, and $\beta,\gamma\ge0$ are chosen on a validation split (Section~\ref{sec:setup}).
Setting $\beta=\gamma=0$ gives the visual-only baseline.
All priors are estimated from pairs of training images of different cameras, with the earlier image of a pair taken as the origin camera.

\emph{Repository form.}
$\cam = \log\big(P(c_g\mid c_q)+\epsilon\big)$, with $P$ the empirical camera-transition matrix of positive (same-identity) training pairs and $\epsilon=10^{-8}$, so any same-camera pair, which never occurs among training positives, receives $\log\epsilon\approx-18.4$.
$\dt = \log\big(P(b\mid c_q,c_g,y{=}1)+\epsilon\big)$ with 8 gap bins.
Both terms are defined for every pair, and a same-camera pair receives $\log\epsilon$ in both.

\emph{Same-camera penalty only.}
$\cam = \log\epsilon$ if $c_q=c_g$ and $0$ otherwise, and $\dt=0$.
It has a single free parameter $\beta$ and uses no training data, and it encodes exactly the protocol fact above: every same-camera gallery image is a distractor that the penalty pushes out of the top of the list.

\emph{Cross-camera only.}
The learned terms of the repository form, set to $0$ for same-camera pairs, so the penalty cannot act.

\emph{Likelihood-ratio (LR) prior.}
For a cross-camera pair in the cell $x=(c_q,c_g,b)$, with $b$ one of 15 logarithmically spaced gap bins (edges 25 to 204{,}800 frames),
\begin{equation}
\lr(x)=\log\frac{P(x\mid y{=}1)}{P(x)} = \log p(y{=}1\mid x)-\log p(y{=}1),
\label{eq:lr}
\end{equation}
where both distributions are estimated by counts with Laplace smoothing ($\alpha=0.5$) from the training images: the numerator from all same-identity cross-camera pairs and the denominator from all cross-camera pairs.
Equation~\eqref{eq:lr} shows that $\lr$ is the log of the conditional match probability of a cell up to a constant, the quantity estimated in \cite{wang2018spatial}.
Used as $\dt$ it is $0$ for same-camera pairs.
With $\cam$ set to the same-camera penalty the fusion is \emph{LR + penalty}; with $\beta=0$ it is \emph{LR, cross-camera only}.

\emph{Time-shuffled controls.}
To test whether a gain depends on the link between time and identity, each prior is rebuilt after permuting the pseudo-time stamps among the images of each camera in the training set.
This keeps the marginal distribution of times in each camera and the cell structure, and breaks only the association of time with identity.
At test time the true time stamps are used, so any remaining gain cannot come from time as evidence about identity.

\subsection{System Architecture}
\begin{figure*}[t]
\centering
\includegraphics[width=0.98\textwidth]{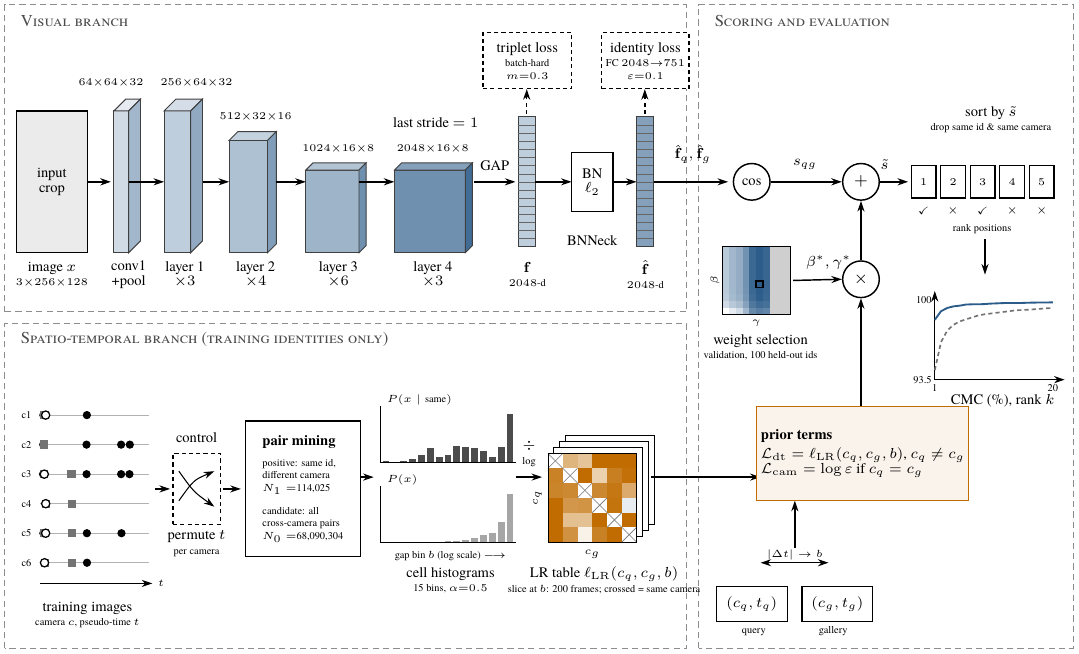}
\caption{Architecture of the audited re-ranker. \emph{Visual branch:} a ResNet-50 whose stages are drawn as blocks (width $\propto$ channels, height $\propto$ spatial size; sizes for a $3{\times}256{\times}128$ input), with the last stride set to 1, global average pooling, a batch-normalisation neck and an L2-normalised 2048-d feature; the triplet and identity losses (dashed arrows) act during training only. \emph{Spatio-temporal branch:} training images are placed on a camera--time plane, optionally with time stamps permuted within each camera (dashed control), positive and candidate pairs are mined, two cell histograms are counted and their log ratio is stored as a $6{\times}6{\times}15$ table. \emph{Scoring:} for each query--gallery pair the cosine score is added to the prior terms read from the table and scaled by the weights chosen on a held-out validation split; the gallery is ranked after the protocol filter and scored by CMC and mAP. The glyphs are drawn from real results: the two histograms (all camera pairs pooled, \NPosShort\ positive and \NAllShort\ candidate pairs), the front slice of the LR table (200-frame bin; crossed cells are same-camera pairs), the validation landscape of $(\beta,\gamma)$ for seed 0 with the chosen cell framed, the CMC curves (mean of three seeds, visual only dashed) and the training images of three identities in one sequence. The numbered boxes and marks are illustrative.}
\label{fig:arch}
\end{figure*}
Fig.~\ref{fig:arch} shows the system.
The \emph{visual branch} is a ResNet-50 \cite{he2015deep} initialised from ImageNet \cite{russakovsky2014imagenet}, with the stride of the last stage set to 1 so that the final map is $2048{\times}16{\times}8$ for a $256{\times}128$ input; global average pooling gives $\mathbf{f}\in\mathbb{R}^{2048}$, which feeds the triplet loss \cite{hermans2017defense}, and a batch-normalisation neck \cite{ioffe2015batch} gives the feature that feeds the label-smoothed identity loss \cite{szegedy2015rethinking} and is, after L2 normalisation, the retrieval feature $\fh$.
The \emph{spatio-temporal branch} sees only the camera, sequence and frame numbers of the training images and never their pixels: it converts them to pseudo-time, mines positive and candidate pairs, accumulates the two cell histograms and stores $\lr$.
At inference the \emph{pair metadata} of a query and a gallery image selects one cell, the \emph{prior terms} are read from the table (or set by the same-camera rule), and the \emph{score fusion} of Eq.~\eqref{eq:fuse} produces the ranking.
The \emph{weight selection} is performed once per snapshot on an identity-disjoint validation split.
The dashed \emph{control} randomises the pseudo-time of the training images before pair mining.

\begin{algorithm}[t]
\caption{Audited re-ranking with a likelihood-ratio prior}
\label{alg:rerank}
\begin{algorithmic}[1]
\REQUIRE training images with identity, camera, sequence, frame; trained encoder; query set $Q$; gallery set $G$
\STATE compute $t$ for all images; \textbf{if} control \textbf{then} permute $t$ within each camera
\STATE $N_1\leftarrow$ counts of cells $x$ over same-identity cross-camera training pairs
\STATE $N_0\leftarrow$ counts of cells $x$ over all cross-camera training pairs
\STATE $\lr(x)\leftarrow\log\hat{P}(x\mid y{=}1)-\log\hat{P}(x)$ for every cross-camera cell, with $\hat{P}$ the $\alpha$-smoothed normalised counts of $N_1$ and $N_0$
\STATE choose $(\beta^{\ast},\gamma^{\ast})$ maximising validation mAP over the grid
\FORALL{$q\in Q$}
  \STATE $\fh_q\leftarrow$ encode$(q)$
  \FORALL{$g\in G$ \textbf{with} not (same identity \textbf{and} same camera)}
    \STATE $\cam\leftarrow\log\epsilon\cdot[c_q{=}c_g]$; $\dt\leftarrow\lr(c_q,c_g,b(|t_q-t_g|))\cdot[c_q{\neq}c_g]$
    \STATE $\tilde{s}_{qg}\leftarrow\fh_q^{\top}\fh_g+\beta^{\ast}\cam+\gamma^{\ast}\dt$
  \ENDFOR
  \STATE rank $G$ by $\tilde{s}_{q\cdot}$ and accumulate AP and CMC
\ENDFOR
\end{algorithmic}
\end{algorithm}
Algorithm~\ref{alg:rerank} summarises the procedure.
The normalisation in line~4 is over the cross-camera cells.
Fusion needs one table look-up per query--gallery pair on top of the dot product of the visual score, so its cost is $O(|Q||G|)$, and the table has only $6{\times}6{\times}15$ entries.

%% file: sections/setup.tex
\section{Experimental Setup}
\label{sec:setup}
\subsection{Data and Protocol}
We use Market-1501 \cite{zheng2015scalable}: 751 training identities (12{,}936 images), 3{,}368 query images and \NGalleryTest\ gallery images from 6 cameras; junk images (identity $-1$) are removed from the gallery and the 2{,}798 distractor images (identity 0) are kept as negatives.
We follow the standard protocol of Section~\ref{sec:method} and report Rank-1, Rank-5, Rank-10 and mAP.
Other benchmarks (for example MSMT17 \cite{wei2018person}) are outside the scope of this study; Section~\ref{sec:discussion} discusses what that implies.

\subsection{Visual Baseline}
The backbone is the ResNet-50 of Fig.~\ref{fig:arch}, trained with a label-smoothed cross-entropy and a batch-hard triplet loss on batches of identities and images, with Adam \cite{kingma2015adam} and a step schedule with warm-up.
Table~\ref{tab:hparams} lists all settings.
Market-1501 images are $128{\times}64$ and are resized to $256{\times}128$; training uses flipping, padding with random cropping and random erasing \cite{zhong2017random}, and bfloat16 autocast \cite{micikevicius2017mixed,kalamkar2019study} in PyTorch \cite{paszke2019pytorch}.
Test features are saved at epochs 0 (the ImageNet network with an untrained neck), 5, 10, 20, 40, 70 and 120.
Snapshots of one run share their weights and schedule, so they are correlated measurements of baseline strength and not independent models.
\begin{table}[!t]
\caption{Training and selection settings.}
\label{tab:hparams}
\centering
\footnotesize\setlength{\tabcolsep}{4pt}
\begin{tabular}{@{}p{0.27\columnwidth}p{0.66\columnwidth}@{}}
\toprule
Item & Setting \\ \midrule
Backbone & ResNet-50, ImageNet initialisation, last stride 1, BNNeck \\
Input & $256{\times}128$ (resized from $128{\times}64$) \\
Loss & label-smoothed CE ($0.1$) + batch-hard triplet (margin $0.3$) \\
Batch & $P{\times}K=16{\times}4$ identities $\times$ images, about 200 iterations per epoch \\
Optimiser & Adam, base rate $3.5{\times}10^{-4}$, L2 weight decay $5{\times}10^{-4}$ \\
Schedule & 10-epoch linear warm-up from $0.1\times$; decay by 10 at epochs 40 and 70; 120 epochs \\
Augmentation & flip, pad 10 px + random crop, random erasing ($p{=}0.5$, area 2--40\,\%) \\
Precision & bfloat16 autocast \\
Seeds & 0, 1, 2 (training only; split and priors fixed) \\
Validation & 100 held-out identities, \NValQ\ queries, \NValG\ gallery images \\
Weight grid & $\beta$ and $\gamma$ each from 0, 0.002, 0.005, 0.01, 0.02, 0.03, 0.05, 0.1, 0.2, 0.5 \\
LR table & 15 log gap bins, $\alpha=0.5$, $\varepsilon=10^{-8}$ \\ \bottomrule
\end{tabular}
\end{table}

\subsection{Choosing the Fusion Weights}
The fusion weights are never chosen on the test set.
For each seed and snapshot we train a second network on 651 training identities and hold out 100 identities as a validation set (one random query per identity and camera, giving \NValQ\ queries and \NValG\ gallery images).
$(\beta,\gamma)$ is chosen from the grid of Table~\ref{tab:hparams} by validation mAP, with priors built from the 651 identities.
The chosen weights are applied to the test features of the network trained on all 751 identities, with priors built from all 751 identities.
Free parameters per configuration are listed in Table~\ref{tab:main}; a parameter that a configuration does not use is fixed at $0$.
We report the gain over the visual-only baseline of the same network.

\subsection{Statistics}
Tables give the mean and standard deviation over the three seeds; the per-seed values and the selected weights are in Appendix~\ref{app:seed}.
Confidence intervals come from a paired bootstrap over the \NQuery\ test queries (\NBoot\ resamples) of the per-query average-precision difference averaged over seeds; they capture query sampling and not the variation of training.
The pair-level diagnostics of Section~\ref{sec:results} use all cross-camera pairs among the validation identities.
One consumer GPU (\TimeGpu) was used; a training run takes about 15 minutes.

%% file: sections/results.tex
\section{Results}
\label{sec:results}
\subsection{Baseline and Overall Accuracy}
At epoch 120 the visual-only baseline reaches Rank-1 \BaseRonePm\,\% and mAP \BaseMapPm\,\% on the test set (mean $\pm$ standard deviation over three seeds), close to the 94.5\,\% and 85.9\,\% reported in \cite{luo2019bag} for a related recipe, and far above the 46.3\,\% and 23.2\,\% of the project we started from, whose run with a batch-normalisation neck and triplet loss had stalled at the chance-level training loss.
Table~\ref{tab:full} gives the full metrics of the four re-rankers; all gains below are measured against this baseline.
\begin{table}[t]
\caption{Test accuracy (\%) at epoch 120, mean $\pm$ standard deviation over three seeds, with the fusion weights chosen on the validation split.}
\label{tab:full}
\centering
\scriptsize\setlength{\tabcolsep}{2.2pt}
\input{tables/tab_full}
\end{table}
The LR prior with the same-camera penalty raises Rank-1 from \BaseRone\,\% to \AbsLrpOne\,\%, Rank-5 from \CmcBaseFive\,\% to \CmcLrpFive\,\% and Rank-10 from \CmcBaseTen\,\% to \CmcLrpTen\,\%.

\subsection{How Much Is Protocol, How Much Is Learned (Q1, Q2)}
\begin{figure}[t]
\centering
\includegraphics[width=\columnwidth]{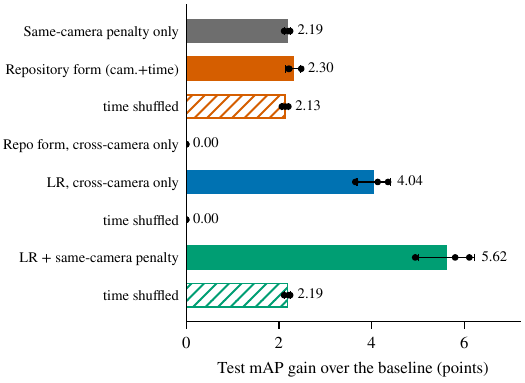}
\caption{Test mAP gain over the baseline at epoch 120. Bars are means, dots are the three seeds, whiskers one standard deviation; hatched bars rebuild the prior from shuffled time stamps.}
\label{fig:decomp}
\end{figure}

Table~\ref{tab:main} and Fig.~\ref{fig:decomp} give the gains in mAP and Rank-1.
\begin{table}[t]
\caption{Test gain over the visual-only baseline at epoch 120 (points), mean $\pm$ standard deviation over three seeds. ``Free'' is the number of weights chosen on validation.}
\label{tab:main}
\centering
\footnotesize\setlength{\tabcolsep}{3pt}
\input{tables/tab_main}
\end{table}

\emph{The repository form is mostly the protocol.}
It adds \GorigMapM\ mAP points (95\,\% CI \Ciorig) and \GorigOneM\ Rank-1 points.
A same-camera penalty with one weight and no training data adds \GscMapM\ points (\Cisc), so \ProtocolShare\,\% of the combined gain of the best prior below is available without learning anything.
The validation split gave the time term of the repository form a weight of \OrigGammaSel\ in the three seeds, and shuffling the time stamps changes the gain only from \GorigMapM\ to \GorigshMapM.
When the same-camera effect is removed (``cross-camera only''), the validation split selects $\beta=\gamma=0$ in every seed: the learned camera and time terms of that form do not help on their own.

\emph{A likelihood-ratio prior adds learned information.}
The LR prior adds \GlrMapM\ points (\Cilr) without any same-camera term, and \GlrpMapM\ points (\Cilrp; Rank-1 \GlrpOneM) with it.
The gain of LR + penalty over the penalty alone is \TimeBeyondProtocolM\ points (\CilrpMinusSc), and over the repository form \BmlrpMinusOrig\ points (\CilrpMinusOrig).

\emph{The extra gain is tied to identity.}
With shuffled time stamps the cross-camera LR prior adds \GlrshMapM\ points and LR + penalty adds \GlrpshMapM, the same as the penalty alone (\GscMapM).
The difference between the unshuffled and shuffled cross-camera LR prior is \BmlrMinusSh\ points (\CilrMinusSh).

\subsection{Why the Repository's Time Term Is Weak}
Table~\ref{tab:auc} measures how well each score separates same-identity from different-identity cross-camera pairs among the 100 validation identities (\DiagPairs\ pairs, \DiagPos\ of them matches, \DiagPosPct\,\%), with priors built from the other 651 identities.
\begin{table}[t]
\caption{AUC of single scores for telling matches from non-matches among cross-camera validation pairs. The second row uses only pairs whose images are in the same sequence.}
\label{tab:auc}
\centering
\footnotesize\setlength{\tabcolsep}{3pt}
\begin{tabular}{@{}lcccccc@{}}
\toprule
 & $-|\Delta t|$ & same seq. & $\cam$ & $\dt$ (repo) & both & LR \\ \midrule
All pairs & \AucRaw & \AucSeq & \AucCam & \AucDelta & \AucBoth & \AucLR \\
Same sequence & \AucWithinRaw & -- & -- & \AucWithinDelta & -- & \AucWithinLR \\ \bottomrule
\end{tabular}
\end{table}
The repository's time term gives an AUC of \AucDelta, close to chance, while the raw gap gives \AucRaw\ and the LR prior \AucLR.
Within one sequence the raw gap reaches \AucWithinRaw\ and the LR prior \AucWithinLR\ but the repository term only \AucWithinDelta.
The raw time gap is clearly informative; the repository's term wastes it, as Eq.~\eqref{eq:bayes} predicts for a term that is not normalised by the candidate distribution.
Matches are also far more likely to share a sequence with the query than non-matches (\ShareSeqPos\,\% against \ShareSeqNeg\,\% of pairs), which matters below.

\subsection{The Gain Shrinks as the Baseline Improves (Q3)}
\begin{figure}[t]
\centering
\includegraphics[width=\columnwidth]{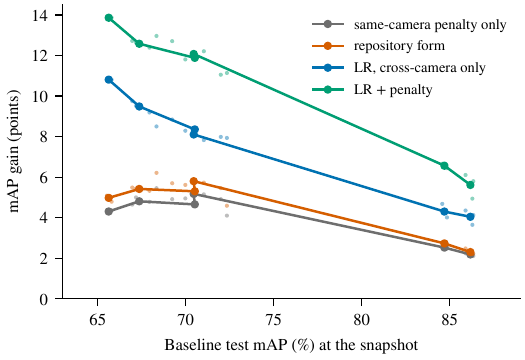}
\caption{Test mAP gain against the baseline mAP of the same snapshot (epochs 5 to 120). Lines join the seed means, dots are single seeds.}
\label{fig:strength}
\end{figure}
\begin{table}[t]
\caption{Baseline accuracy (\%) and mean mAP gain (points) over the three seeds at each snapshot. Pen.: same-camera penalty only; Repo: repository form; LR-x: LR prior, cross-camera only; LR+P: LR prior with the penalty. At epoch 0 the backbone is untrained.}
\label{tab:epochs}
\centering
\footnotesize\setlength{\tabcolsep}{4pt}
\input{tables/tab_epochs}
\end{table}
Fig.~\ref{fig:strength} and Table~\ref{tab:epochs} follow the gain along training.
At epoch 20 the baseline mAP is \EpTwentyBase\,\% and the gains are \EpTwentysc\ (penalty only), \EpTwentyorig\ (repository form), \EpTwentylr\ (LR, cross-camera) and \EpTwentylrp\ points (LR + penalty); at epoch 120 (baseline \BaseMap\,\%) they are \GscMapM, \GorigMapM, \GlrMapM\ and \GlrpMapM.
Every prior helps less on the stronger model, and the penalty and the repository form are nearly indistinguishable at all snapshots.
At epoch 0 the untrained backbone has an mAP of \EpZeroBase\,\% and the LR prior alone adds \EpZerolr\ points, so the prior matters most when appearance is least informative.
The snapshots come from one run per seed, so this trend is evidence about training progress and not about independent models of different strength.

\subsection{Where the Gain Appears (Q4)}
\label{sec:where}
\begin{figure}[t]
\centering
\includegraphics[width=\columnwidth]{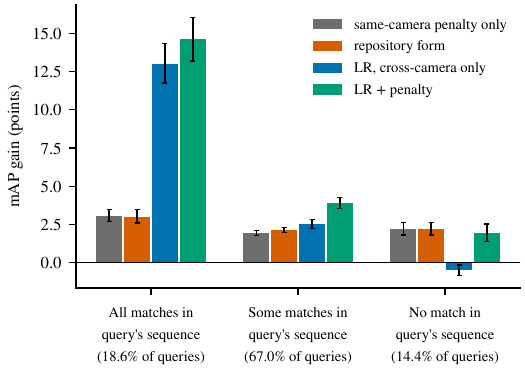}
\caption{Test mAP gain by whether the true matches of a query lie in the query's video sequence (determined from the labels, so descriptive). Whiskers are 95\,\% bootstrap intervals over queries.}
\label{fig:strata}
\end{figure}
We split the test queries by the share of their true matches (images of the same identity in other cameras) that lie in the same sequence as the query: all (\StrShareall\,\% of queries), some (\StrSharesome\,\%) or none (\StrSharenone\,\%).
The baseline mAP of these groups is \StrBaseall, \StrBasesome\ and \StrBasenone\,\%.
For queries whose matches are all in the query's sequence, the penalty adds \Strallsc, the repository form \Strallorig, the cross-camera LR prior \Stralllr\ and LR + penalty \Stralllrp\ points.
For queries with some matches in the sequence the corresponding gains are \Strsomesc, \Strsomeorig, \Strsomelr\ and \Strsomelrp, and for queries with no match in the sequence they are \Strnonesc, \Strnoneorig, \Strnonelr\ (\StrnonelrCi) and \Strnonelrp.
The identity-linked part of the LR gain is therefore concentrated on queries whose matches are close in time, and the cross-camera LR prior slightly hurts queries whose matches are far away.

\begin{table}[t]
\caption{Test mAP by query camera at epoch 120: baseline average precision (\%) and gain (points) of the penalty (Pen.), the repository form (Repo) and the LR prior with the penalty (LR+P), averaged over seeds.}
\label{tab:cam}
\centering
\footnotesize\setlength{\tabcolsep}{4pt}
\input{tables/tab_cam}
\end{table}
Table~\ref{tab:cam} breaks the gain down by query camera.
The penalty and the repository form give similar gains in every camera (between \CamPRMin\ and \CamPRMax\ points), whereas the LR gain varies from \CamGainMin\ to \CamGainMax\ points.
It is largest for camera 4, which has the lowest baseline precision (\CamLowBaseAP\,\%), and smallest for camera 3, which has the highest (\CamHighBaseAP\,\%); with six cameras this is suggestive and not conclusive evidence that the learned prior helps most where appearance is weakest.

\subsection{Qualitative Analysis}
\begin{figure*}[t]
\centering
\includegraphics[width=\textwidth]{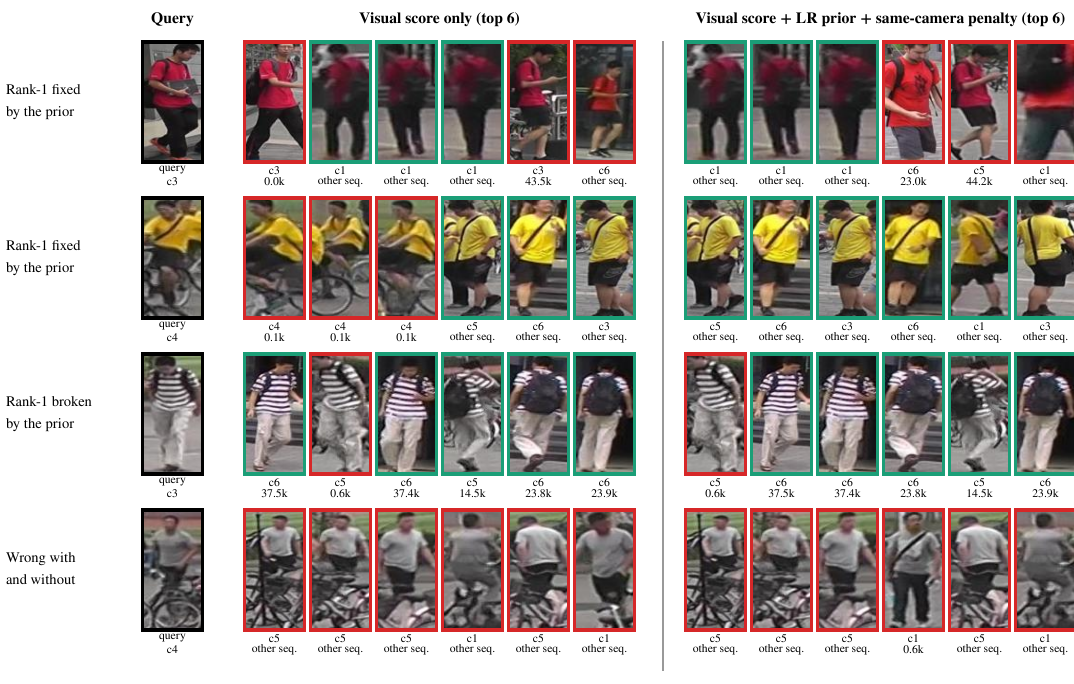}
\caption{Retrieval examples (seed 0, epoch 120). Green frames are true matches and red frames are wrong identities; below each tile the gallery camera and the time gap to the query (in thousand frames, or ``other seq.'' when the images are in different sequences). One query was drawn at random, with a fixed seed, from each category; none was hand-picked.}
\label{fig:qual}
\end{figure*}
Fig.~\ref{fig:qual} shows retrieval lists with and without the LR prior and the penalty.
Over the \NQuery\ queries (seed 0) the first-ranked image changes from wrong to right for \Qualimproved\ queries (\QualPctimproved\,\%), from right to wrong for \Qualworsened\ (\QualPctworsened\,\%), is right in both for \Qualbothcorrect\ and wrong in both for \Qualbothwrong\ (\QualPctbothwrong\,\%).
In the first two rows the visual score ranks images of other people from the query's own camera, taken within a fraction of a second, above the true matches; the penalty removes them and the LR prior favours matches in plausible camera pairs.
In the third row the prior breaks a correct first rank: a different person seen in camera 5 only 600 frames after the query is time-close and receives a high likelihood ratio, which pushes the true match, seen in camera 6, to second place.
In the fourth row the visual score and the prior both fail, since the wearer of similar grey clothes is separated from the query by neither camera nor time.
These failures show what a time prior cannot do: it ranks \emph{plausibility}, not identity.

\begin{figure*}[t]
\centering
\includegraphics[width=\textwidth]{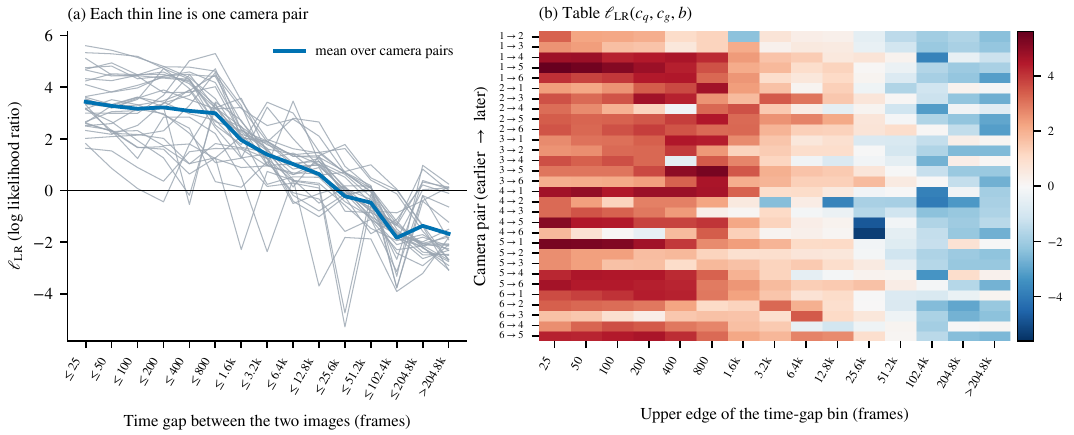}
\caption{The learned likelihood-ratio table built from all 751 training identities. (a) Log likelihood ratio against the time gap for each of the 30 ordered camera pairs (thin lines) and their mean (bold). (b) The same table as a heat map; red cells favour a match, blue cells disfavour it.}
\label{fig:prior}
\end{figure*}
Fig.~\ref{fig:prior} shows what the prior has learned.
For gaps up to 800 frames the mean log likelihood ratio is about 3 to 3.5, i.e.\ a pair at such a gap is roughly 20 to 30 times more likely to be a match than a random cross-camera pair, and the mean falls below zero beyond a gap of about 25{,}000 frames.
For gaps above 100{,}000 frames, which correspond to images of different sequences, it is clearly negative (about $-1.5$).
The spread across camera pairs is considerable, which is why the camera pair is part of the cell.

\subsection{Sensitivity and Robustness}
\begin{table}[t]
\caption{Sensitivity of the mAP gain (points, mean $\pm$ standard deviation over three seeds) to the number of gap bins and to the smoothing constant $\alpha$ of the LR table. Weights are re-selected on validation for each design.}
\label{tab:sens}
\centering
\footnotesize\setlength{\tabcolsep}{4pt}
\input{tables/tab_sens}
\end{table}
\begin{figure*}[t]
\centering
\includegraphics[width=\textwidth]{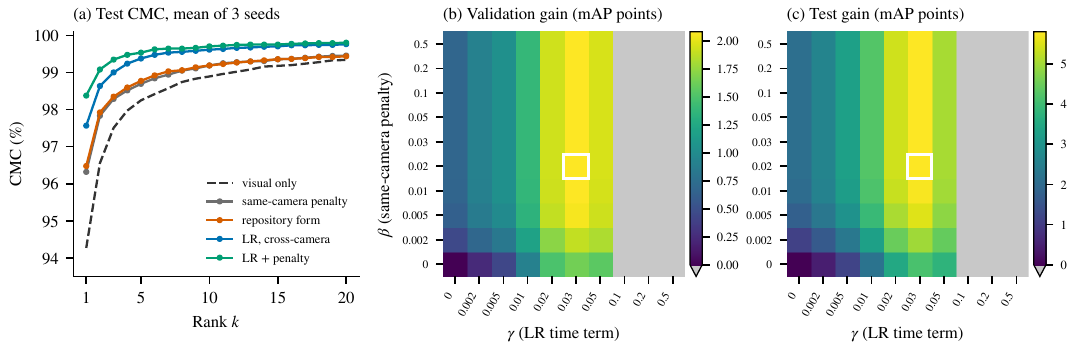}
\caption{(a) CMC curves at epoch 120 (mean of three seeds). (b, c) Gain of the LR prior with the penalty over a grid of weights on the validation split and on the test set (seed 0); the white frame marks the weights chosen on validation, grey cells have a negative gain, and the colour scales differ between (b) and (c).}
\label{fig:cmc}
\end{figure*}
Table~\ref{tab:sens} varies the design of the LR table.
The gain of LR + penalty stays between \SensMin\ and \SensMax\ points for 8, 15 and 28 gap bins and for $\alpha\in\{0.1,0.5,2\}$, so the conclusions do not hinge on these choices.
Fig.~\ref{fig:cmc}(a) shows the CMC curves: the priors lift the curve mostly at ranks 1 to 5 and converge to the baseline by rank 20.
Fig.~\ref{fig:cmc}(b, c) shows the weight landscape for seed 0.
The validation gain is much smaller than the test gain (at most \LandValMax\ against \LandTestMax\ points), which is consistent with a validation gallery of only \NValG\ images against \NGalleryTest\ and therefore far fewer distractors for the penalty and the prior to remove; yet the weights chosen on validation ($\beta=\LandBeta$, $\gamma=\LandGamma$) reach \LandTestAtSel\ points on the test set, the best value of the grid.
The surface is flat in $\beta$ once $\beta\ge0.01$ and has a clear ridge in $\gamma$ at 0.02 to 0.05; time weights of $\gamma\ge0.1$ are harmful (the test gain is negative, grey cells), so the validation split is needed to avoid over-weighting the prior.

\subsection{Computational Cost}
The re-ranker is cheap.
On a \TimeGpu\ GPU, for all $3{,}368\times15{,}913$ query--gallery pairs, the cosine similarity matrix takes \TimeSim\,ms, reading the prior terms from the table takes \TimeTerms\,ms and the fusion of Eq.~\eqref{eq:fuse} takes \TimeFuse\,ms, so the prior adds about \TimeFusionTotal\,ms to the visual score; the full evaluation, including sorting, takes \TimeEval\,ms.
Building the table from the 12{,}936 training images takes \TimeTable\,ms.
These are median wall-clock times over seven runs (three for the table and the evaluation) on cached features, excluding the visual encoder, whose cost is the same with or without the prior.

%% file: tables/tab_full.tex
\begin{tabular}{@{}lcccc@{}}
\toprule
Method & Rank-1 & Rank-5 & Rank-10 & mAP \\ \midrule
Visual score only & 94.27\,$\pm$\,0.05 & 98.25\,$\pm$\,0.00 & 98.89\,$\pm$\,0.11 & 86.18\,$\pm$\,0.24 \\
\midrule
Same-camera penalty only & 96.32\,$\pm$\,0.13 & 98.69\,$\pm$\,0.10 & 99.19\,$\pm$\,0.03 & 88.37\,$\pm$\,0.19 \\
Repository form (camera+time) & 96.48\,$\pm$\,0.11 & 98.77\,$\pm$\,0.03 & 99.19\,$\pm$\,0.05 & 88.48\,$\pm$\,0.09 \\
LR prior, cross-camera only & 97.57\,$\pm$\,0.19 & 99.38\,$\pm$\,0.08 & 99.61\,$\pm$\,0.05 & 90.23\,$\pm$\,0.27 \\
LR prior + same-camera penalty & 98.38\,$\pm$\,0.20 & 99.53\,$\pm$\,0.03 & 99.70\,$\pm$\,0.05 & 91.80\,$\pm$\,0.49 \\
\bottomrule
\end{tabular}

%% file: tables/tab_main.tex
\begin{tabular}{@{}lccc@{}}
\toprule
Prior & Free & $\Delta$mAP & $\Delta$Rank-1 \\ \midrule
Same-camera penalty only & 1 & 2.19\,$\pm$\,0.07 & 2.05\,$\pm$\,0.11 \\
Repository form (camera+time) & 2 & 2.30\,$\pm$\,0.15 & 2.21\,$\pm$\,0.09 \\
\quad time shuffled & 2 & 2.13\,$\pm$\,0.07 & 2.02\,$\pm$\,0.13 \\
Repository form, cross-camera only & 2 & 0.00\,$\pm$\,0.00 & 0.00\,$\pm$\,0.00 \\
LR prior, cross-camera only & 1 & 4.04\,$\pm$\,0.36 & 3.30\,$\pm$\,0.18 \\
\quad time shuffled & 1 & 0.00\,$\pm$\,0.00 & 0.00\,$\pm$\,0.00 \\
LR prior + same-camera penalty & 2 & 5.62\,$\pm$\,0.60 & 4.11\,$\pm$\,0.15 \\
\quad time shuffled & 2 & 2.19\,$\pm$\,0.07 & 2.06\,$\pm$\,0.12 \\
\bottomrule
\end{tabular}

%% file: tables/tab_epochs.tex
\begin{tabular}{@{}rcccccc@{}}
\toprule
Epoch & mAP & R1 & Pen. & Repo & LR-x & LR+P \\ \midrule
0 & 2.7 & 8.9 & +1.4 & +1.6 & +13.7 & +14.0 \\
5 & 65.6 & 83.9 & +4.3 & +5.0 & +10.8 & +13.9 \\
10 & 67.4 & 85.3 & +4.8 & +5.4 & +9.5 & +12.6 \\
20 & 70.5 & 87.0 & +4.7 & +5.3 & +8.4 & +11.9 \\
40 & 70.5 & 87.3 & +5.2 & +5.8 & +8.1 & +12.1 \\
70 & 84.7 & 93.7 & +2.5 & +2.7 & +4.3 & +6.6 \\
120 & 86.2 & 94.3 & +2.2 & +2.3 & +4.0 & +5.6 \\
\bottomrule
\end{tabular}

%% file: tables/tab_cam.tex
\begin{tabular}{@{}cccccc@{}}
\toprule
Query cam. & Queries & Base AP & Pen. & Repo & LR + P \\ \midrule
1 & 668 & 85.6 & +2.1 & +2.2 & +4.3 \\
2 & 530 & 88.5 & +1.7 & +2.1 & +5.6 \\
3 & 717 & 89.7 & +1.6 & +1.8 & +4.1 \\
4 & 248 & 80.2 & +2.0 & +1.4 & +7.8 \\
5 & 622 & 83.7 & +2.3 & +2.4 & +6.7 \\
6 & 583 & 85.6 & +3.4 & +3.6 & +7.0 \\
\bottomrule
\end{tabular}

%% file: tables/tab_sens.tex
\begin{tabular}{@{}lcc@{}}
\toprule
LR table design & LR, cross-camera & LR + penalty \\ \midrule
8 bins, $\alpha=0.5$ & 3.83\,$\pm$\,0.16 & 5.44\,$\pm$\,0.37 \\
15 bins, $\alpha=0.5$ (default) & 4.04\,$\pm$\,0.36 & 5.62\,$\pm$\,0.61 \\
28 bins, $\alpha=0.5$ & 4.16\,$\pm$\,0.15 & 5.94\,$\pm$\,0.60 \\
15 bins, $\alpha=0.1$ & 4.03\,$\pm$\,0.37 & 5.60\,$\pm$\,0.61 \\
15 bins, $\alpha=2$ & 4.06\,$\pm$\,0.35 & 5.63\,$\pm$\,0.60 \\
\bottomrule
\end{tabular}

%% file: sections/discussion.tex
\section{Discussion}
\label{sec:discussion}
\subsection{What the Findings Say}
The four questions of Section~I have short answers on this dataset.
\emph{Q1:} for an additive log-prior fusion learned in the repository's form, the same-camera effect accounts for nearly all of the gain, and the learned time term is almost unused because it is not normalised by the candidate distribution.
\emph{Q2:} a likelihood-ratio prior adds about \TimeBeyondProtocolM\ mAP points that vanish when time and identity are decoupled, so they are evidence about identity and not an artefact of reshaping the scores.
\emph{Q3:} all gains shrink as the visual model improves, from \EpTwentylrp\ to \GlrpMapM\ points for the best prior.
\emph{Q4:} the identity-linked gain is concentrated on queries whose matches are close in time and on some cameras, and the prior ranks plausibility and not identity, so it can promote a time-close wrong person.

\subsection{Recommendations for Reporting}
From these findings we suggest that a re-ID paper that adds camera or time information should:
\begin{enumerate}
\item report a same-camera-penalty control next to the visual baseline, since it can account for most of a reported gain;
\item report a time-shuffled control, which is cheap and shows whether the prior carries identity information at all;
\item state how the weights were chosen and use data disjoint from the test identities;
\item compare against a strong visual baseline and show how the gain changes with its strength, as the gain falls with baseline quality (Fig.~\ref{fig:strength});
\item report where the gain occurs, for instance by sequence overlap or by camera, because aggregate gains hide large differences between queries.
\end{enumerate}
These steps are inexpensive: on cached features, one configuration needs about \TimeFusionTotal\,ms for the prior and the fusion and \TimeEval\,ms for the evaluation (Section~\ref{sec:results}).

\subsection{Limitations and Threats to Validity}
\emph{One dataset.}
All results are on Market-1501; other benchmarks with different camera layouts and recording spans may behave differently, and we did not test them.
The finding that gains concentrate on matches in the same sequence is a property of how this dataset was recorded and should not be extrapolated.
\emph{One baseline family and three seeds.}
We use one backbone and loss family; the standard deviations are over three training seeds, and the bootstrap intervals cover query sampling only.
\emph{Same session.}
Training and test identities were recorded in the same session, so the priors are tested across identities but not across days or camera networks; a prior estimated in one network need not transfer to another, in line with the broader evidence on distribution shift \cite{koh2021wilds}.
\emph{Pseudo-time.}
The time stamp is built from sequence and frame numbers with the offset of the original repository; we did not check it against real clock time.
\emph{Selection.}
The fusion weights come from a validation split of 100 identities whose gallery is small; the validation gain understates the test gain (Fig.~\ref{fig:cmc}), although on seed 0 the weights chosen there reached the best value of the grid on the test set, and in a diagnostic run the weights tuned directly on the test set gave gains within 0.03 points of the validated ones.
\emph{Labels in the analysis.}
The split by shared sequence and the qualitative categories use test labels and are descriptive, not part of the method.
\emph{Scope of the comparison.}
We did not re-implement the models of \cite{wang2018spatial,ren2021learning,lv2018unsupervised}, whose smoothing, joint metrics or training differ from our additive fusion, and we make no claim about their reported numbers.
\emph{Independence assumption.}
Eq.~\eqref{eq:bayes} assumes that the visual and the spatio-temporal evidence are conditionally independent given identity; this is plausible but untested.

\subsection{Directions for Future Work}
Four extensions follow from the limitations above.
First, repeating the audit on benchmarks with different recording spans and camera layouts, for example MSMT17 \cite{wei2018person} and a spatio-temporal benchmark such as the one of \cite{shu2021large}, would show whether the concentration of the gain on same-sequence matches is a property of Market-1501 or of the method.
Second, a prior estimated on one camera network should be tested on another, to measure how much of the identity-linked gain survives the shift between networks \cite{koh2021wilds}, possibly with the adaptation methods reviewed in Section~\ref{sec:related}.
Third, the learned prior could be combined with the stronger visual models cited above, and with list-level re-ranking \cite{zhong2017re,sarfraz2018pose}, to see whether the gains add up or overlap.
Fourth, the weights of Eq.~\eqref{eq:fuse} are selected on a small validation split; a calibrated visual score \cite{guo2017calibration} would allow the weights to be derived from Eq.~\eqref{eq:bayes} instead of being searched.

\subsection{Ethical Considerations}
Person re-identification supports surveillance applications.
The benchmark gains analysed here should not be read as evidence that a system is ready for deployment, and a prior that exploits camera and time of capture additionally depends on retaining location and time metadata of individuals.
Market-1501 consists of low-resolution images of pedestrians collected for research; we show a small number of them in Fig.~\ref{fig:qual} only to illustrate failure and success cases, and we do not redistribute the dataset.

%% file: sections/conclusion.tex
\section{Conclusion}
\label{sec:conclusion}
On a strong ResNet-50 baseline (mAP \BaseMap\,\%), an additive camera and time prior in the form of our starting repository adds \GorigMapM\ mAP points, and a one-weight same-camera penalty adds \GscMapM: the evaluation protocol alone explains nearly all of it.
A likelihood-ratio prior, which normalises the time term by the candidate distribution as Bayes' rule requires, adds \GlrMapM\ points without the penalty and \GlrpMapM\ with it, and about \TimeBeyondProtocolM\ of these points vanish when the time stamps are shuffled, so they are tied to identity.
The prior corrects the first-ranked match for \QualPctimproved\,\% of the queries and breaks it for \QualPctworsened\,\%, its gain shrinks as the visual model improves, and it concentrates on queries whose matches share the query's sequence.
We recommend that work on spatio-temporal re-ID report a same-camera control, a shuffled-time control and a strong baseline, and that the conclusions of this study be tested on further datasets and on camera networks other than the one used for training.

%% file: sections/appendix.tex
\appendices
\section{Per-Seed Results and Selected Weights}
\label{app:seed}
Table~\ref{tab:seed} gives, for each training seed, the test accuracy at epoch 120 and the weights selected on the validation split.
The selected weights are stable: the LR prior with the penalty uses $\beta^{\ast}\in\{\LrpBetaSel\}$ and $\gamma^{\ast}\in\{\LrpGammaSel\}$ in the three seeds, and the time weight of the repository form is $\{\OrigGammaSel\}$, i.e.\ close to zero.
In every seed LR + penalty exceeds LR cross-camera only, which exceeds the repository form and the penalty, and these last two are within 0.3 points of each other; in seed 2 the validation split even gave the repository form a time weight of zero, so it coincides with the penalty.
\begin{table}[!t]
\caption{Test results at epoch 120 for each seed (\%, gain in points). $\beta^{\ast}$ and $\gamma^{\ast}$ are the weights selected on validation; ``--'' marks the visual-only baseline.}
\label{tab:seed}
\centering
\footnotesize\setlength{\tabcolsep}{4pt}
\input{tables/tab_seed}
\end{table}

%% file: tables/tab_seed.tex
\begin{tabular}{@{}clccccc@{}}
\toprule
Seed & Prior & $\beta^{\ast}$ & $\gamma^{\ast}$ & Rank-1 & mAP & $\Delta$mAP \\ \midrule
0 & visual only & -- & -- & 94.24 & 86.35 & -- \\
 & same-camera penalty & 0.03 & 0 & 96.17 & 88.46 & +2.11 \\
 & repository form & 0.01 & 0.005 & 96.35 & 88.57 & +2.22 \\
 & LR, cross-camera & 0 & 0.03 & 97.36 & 90.48 & +4.13 \\
 & LR + penalty & 0.02 & 0.03 & 98.28 & 92.15 & +5.80 \\
\midrule
1 & visual only & -- & -- & 94.33 & 85.90 & -- \\
 & same-camera penalty & 0.03 & 0 & 96.41 & 88.15 & +2.24 \\
 & repository form & 0.02 & 0.005 & 96.56 & 88.39 & +2.48 \\
 & LR, cross-camera & 0 & 0.03 & 97.62 & 90.26 & +4.35 \\
 & LR + penalty & 0.03 & 0.03 & 98.60 & 92.01 & +6.11 \\
\midrule
2 & visual only & -- & -- & 94.24 & 86.29 & -- \\
 & same-camera penalty & 0.03 & 0 & 96.38 & 88.50 & +2.21 \\
 & repository form & 0.03 & 0 & 96.53 & 88.50 & +2.21 \\
 & LR, cross-camera & 0 & 0.05 & 97.71 & 89.94 & +3.65 \\
 & LR + penalty & 0.03 & 0.05 & 98.25 & 91.23 & +4.94 \\
\bottomrule
\end{tabular}

%% file: main.bbl
\begin{thebibliography}{99}
\setlength{\itemsep}{0pt}
\footnotesize

\bibitem{zheng2015scalable} L. Zheng, L. Shen, L. Tian, S. Wang, J. Wang, and Q. Tian, ``Scalable person re-identification: A benchmark,'' in Proc. IEEE Int. Conf. Comput. Vis. (ICCV), 2015, pp. 1116--1124, doi:10.1109/ICCV.2015.133.
\bibitem{wang2018spatial} G. Wang, J. Lai, P. Huang, and X. Xie, ``Spatial-temporal person re-identification,'' arXiv:1812.03282, 2018 (AAAI 2019).
\bibitem{luo2019bag} H. Luo, Y. Gu, X. Liao, S. Lai, and W. Jiang, ``Bag of tricks and a strong baseline for deep person re-identification,'' in CVPR Workshops, 2019, arXiv:1903.07071.
\bibitem{hermans2017defense} A. Hermans, L. Beyer, and B. Leibe, ``In defense of the triplet loss for person re-identification,'' arXiv:1703.07737, 2017.
\bibitem{zhong2017random} Z. Zhong, L. Zheng, G. Kang, S. Li, and Y. Yang, ``Random erasing data augmentation,'' arXiv:1708.04896, 2017.
\bibitem{he2015deep} K. He, X. Zhang, S. Ren, and J. Sun, ``Deep residual learning for image recognition,'' arXiv:1512.03385, 2015.
\bibitem{kingma2015adam} D. P. Kingma and J. Ba, ``Adam: A method for stochastic optimization,'' in Proc. Int. Conf. Learn. Represent. (ICLR), 2015, arXiv:1412.6980.
\bibitem{szegedy2015rethinking} C. Szegedy, V. Vanhoucke, S. Ioffe, J. Shlens, and Z. Wojna, ``Rethinking the Inception architecture for computer vision,'' arXiv:1512.00567, 2015.
\bibitem{zheng2016past} L. Zheng, Y. Yang, and A. G. Hauptmann, ``Person re-identification: Past, present and future,'' arXiv:1610.02984, 2016.
\bibitem{ye2020deep} M. Ye, J. Shen, G. Lin, T. Xiang, L. Shao, and S. C. H. Hoi, ``Deep learning for person re-identification: A survey and outlook,'' arXiv:2001.04193, 2020 (accepted by IEEE Trans. Pattern Anal. Mach. Intell.).
\bibitem{zhong2017re} Z. Zhong, L. Zheng, D. Cao, and S. Li, ``Re-ranking person re-identification with k-reciprocal encoding,'' in Proc. IEEE/CVF Conf. Comput. Vis. Pattern Recognit. (CVPR), 2017, arXiv:1701.08398.
\bibitem{geirhos2020shortcut} R. Geirhos \emph{et al.}, ``Shortcut learning in deep neural networks,'' \emph{Nature Mach. Intell.}, 2020, arXiv:2004.07780, doi:10.1038/s42256-020-00257-z.
\bibitem{wei2018person} L. Wei, S. Zhang, W. Gao, and Q. Tian, ``Person transfer GAN to bridge domain gap for person re-identification,'' in Proc. IEEE/CVF Conf. Comput. Vis. Pattern Recognit. (CVPR), 2018, arXiv:1711.08565.
\bibitem{sun2017beyond} Y. Sun, L. Zheng, Y. Yang, Q. Tian, and S. Wang, ``Beyond part models: Person retrieval with refined part pooling (and a strong convolutional baseline),'' arXiv:1711.09349, 2017.
\bibitem{wang2018learning} G. Wang, Y. Yuan, X. Chen, J. Li, and X. Zhou, ``Learning discriminative features with multiple granularities for person re-identification,'' in Proc. ACM Int. Conf. Multimedia, 2018, arXiv:1804.01438.
\bibitem{zhou2019omni} K. Zhou, Y. Yang, A. Cavallaro, and T. Xiang, ``Omni-scale feature learning for person re-identification,'' in Proc. IEEE/CVF Int. Conf. Comput. Vis. (ICCV), 2019, arXiv:1905.00953.
\bibitem{zhou2019learning} K. Zhou, Y. Yang, A. Cavallaro, and T. Xiang, ``Learning generalisable omni-scale representations for person re-identification,'' \emph{IEEE Trans. Pattern Anal. Mach. Intell.}, 2021, arXiv:1910.06827.
\bibitem{he2021transreid} S. He, H. Luo, P. Wang, F. Wang, H. Li, and W. Jiang, ``TransReID: Transformer-based object re-identification,'' arXiv:2102.04378, 2021.
\bibitem{dosovitskiy2020image} A. Dosovitskiy \emph{et al.}, ``An image is worth 16x16 words: Transformers for image recognition at scale,'' arXiv:2010.11929, 2020.
\bibitem{schroff2015facenet} F. Schroff, D. Kalenichenko, and J. Philbin, ``FaceNet: A unified embedding for face recognition and clustering,'' in Proc. IEEE/CVF Conf. Comput. Vis. Pattern Recognit. (CVPR), 2015, arXiv:1503.03832.
\bibitem{xiao2016joint} T. Xiao, S. Li, B. Wang, L. Lin, and X. Wang, ``Joint detection and identification feature learning for person search,'' in Proc. IEEE/CVF Conf. Comput. Vis. Pattern Recognit. (CVPR), 2017, arXiv:1604.01850.
\bibitem{chen2017beyond} W. Chen, X. Chen, J. Zhang, and K. Huang, ``Beyond triplet loss: A deep quadruplet network for person re-identification,'' in Proc. IEEE/CVF Conf. Comput. Vis. Pattern Recognit. (CVPR), 2017, arXiv:1704.01719.
\bibitem{li2018harmonious} W. Li, X. Zhu, and S. Gong, ``Harmonious attention network for person re-identification,'' in Proc. IEEE/CVF Conf. Comput. Vis. Pattern Recognit. (CVPR), 2018, arXiv:1802.08122.
\bibitem{chang2018multi} X. Chang, T. M. Hospedales, and T. Xiang, ``Multi-level factorisation net for person re-identification,'' in Proc. IEEE/CVF Conf. Comput. Vis. Pattern Recognit. (CVPR), 2018, arXiv:1803.09132.
\bibitem{chen2019mixed} B. Chen, W. Deng, and J. Hu, ``Mixed high-order attention network for person re-identification,'' in Proc. IEEE/CVF Int. Conf. Comput. Vis. (ICCV), 2019, arXiv:1908.05819.
\bibitem{fu2018horizontal} Y. Fu \emph{et al.}, ``Horizontal pyramid matching for person re-identification,'' in Proc. AAAI Conf. Artif. Intell., 2019, arXiv:1804.05275.
\bibitem{zheng2018pyramidal} F. Zheng \emph{et al.}, ``Pyramidal person re-identification via multi-loss dynamic training,'' in Proc. IEEE/CVF Conf. Comput. Vis. Pattern Recognit. (CVPR), 2019, arXiv:1810.12193.
\bibitem{quan2019auto} R. Quan, X. Dong, Y. Wu, L. Zhu, and Y. Yang, ``Auto-ReID: Searching for a part-aware ConvNet for person re-identification,'' in Proc. IEEE/CVF Int. Conf. Comput. Vis. (ICCV), 2019, arXiv:1903.09776.
\bibitem{deng2018image} W. Deng, L. Zheng, Q. Ye, G. Kang, Y. Yang, and J. Jiao, ``Image-image domain adaptation with preserved self-similarity and domain-dissimilarity for person re-identification,'' in Proc. IEEE/CVF Conf. Comput. Vis. Pattern Recognit. (CVPR), 2018, arXiv:1711.07027.
\bibitem{zhong2018camera} Z. Zhong, L. Zheng, Z. Zheng, S. Li, and Y. Yang, ``Camera style adaptation for person re-identification,'' in Proc. IEEE/CVF Conf. Comput. Vis. Pattern Recognit. (CVPR), 2018, arXiv:1711.10295.
\bibitem{fan2017unsupervised} H. Fan, L. Zheng, and Y. Yang, ``Unsupervised person re-identification: Clustering and fine-tuning,'' arXiv:1705.10444, 2017.
\bibitem{ge2020mutual} Y. Ge, D. Chen, and H. Li, ``Mutual mean-teaching: Pseudo label refinery for unsupervised domain adaptation on person re-identification,'' in Proc. Int. Conf. Learn. Represent. (ICLR), 2020, arXiv:2001.01526.
\bibitem{ge2020self} Y. Ge, F. Zhu, D. Chen, R. Zhao, and H. Li, ``Self-paced contrastive learning with hybrid memory for domain adaptive object re-ID,'' in Adv. Neural Inf. Process. Syst. (NeurIPS), 2020, arXiv:2006.02713.
\bibitem{dai2021cluster} Z. Dai, G. Wang, W. Yuan, X. Liu, S. Zhu, and P. Tan, ``Cluster contrast for unsupervised person re-identification,'' arXiv:2103.11568, 2021 (ACCV 2022).
\bibitem{luo2021self} H. Luo \emph{et al.}, ``Self-supervised pre-training for transformer-based person re-identification,'' arXiv:2111.12084, 2021.
\bibitem{li2022clip} S. Li, L. Sun, and Q. Li, ``CLIP-ReID: Exploiting vision-language model for image re-identification without concrete text labels,'' in Proc. AAAI Conf. Artif. Intell., 2023, arXiv:2211.13977.
\bibitem{zheng2019joint} Z. Zheng, X. Yang, Z. Yu, L. Zheng, Y. Yang, and J. Kautz, ``Joint discriminative and generative learning for person re-identification,'' in Proc. IEEE/CVF Conf. Comput. Vis. Pattern Recognit. (CVPR), 2019, arXiv:1904.07223.
\bibitem{zhong2019invariance} Z. Zhong, L. Zheng, Z. Luo, S. Li, and Y. Yang, ``Invariance matters: Exemplar memory for domain adaptive person re-identification,'' in Proc. IEEE/CVF Conf. Comput. Vis. Pattern Recognit. (CVPR), 2019, arXiv:1904.01990.
\bibitem{zheng2017unlabeled} Z. Zheng, L. Zheng, and Y. Yang, ``Unlabeled samples generated by GAN improve the person re-identification baseline in vitro,'' in Proc. IEEE/CVF Int. Conf. Comput. Vis. (ICCV), 2017, arXiv:1701.07717.
\bibitem{zheng2016wild} L. Zheng, H. Zhang, S. Sun, M. Chandraker, Y. Yang, and Q. Tian, ``Person re-identification in the wild,'' in Proc. IEEE/CVF Conf. Comput. Vis. Pattern Recognit. (CVPR), 2017, arXiv:1604.02531.
\bibitem{sarfraz2018pose} M. S. Sarfraz, A. Schumann, A. Eberle, and R. Stiefelhagen, ``A pose-sensitive embedding for person re-identification with expanded cross neighborhood re-ranking,'' in Proc. IEEE/CVF Conf. Comput. Vis. Pattern Recognit. (CVPR), 2018, arXiv:1711.10378.
\bibitem{ristani2018features} E. Ristani and C. Tomasi, ``Features for multi-target multi-camera tracking and re-identification,'' in Proc. IEEE/CVF Conf. Comput. Vis. Pattern Recognit. (CVPR), 2018, arXiv:1803.10859.
\bibitem{ioffe2015batch} S. Ioffe and C. Szegedy, ``Batch normalization: Accelerating deep network training by reducing internal covariate shift,'' arXiv:1502.03167, 2015.
\bibitem{russakovsky2014imagenet} O. Russakovsky \emph{et al.}, ``ImageNet large scale visual recognition challenge,'' arXiv:1409.0575, 2014.
\bibitem{muller2019does} R. M{\"u}ller, S. Kornblith, and G. Hinton, ``When does label smoothing help?,'' in Adv. Neural Inf. Process. Syst. (NeurIPS), 2019, arXiv:1906.02629.
\bibitem{wu2017sampling} C.-Y. Wu, R. Manmatha, A. J. Smola, and P. Kr{\"a}henb{\"u}hl, ``Sampling matters in deep embedding learning,'' in Proc. IEEE/CVF Int. Conf. Comput. Vis. (ICCV), 2017, arXiv:1706.07567.
\bibitem{devries2017improved} T. DeVries and G. W. Taylor, ``Improved regularization of convolutional neural networks with cutout,'' arXiv:1708.04552, 2017.
\bibitem{micikevicius2017mixed} P. Micikevicius \emph{et al.}, ``Mixed precision training,'' in Proc. Int. Conf. Learn. Represent. (ICLR), 2018, arXiv:1710.03740.
\bibitem{paszke2019pytorch} A. Paszke \emph{et al.}, ``PyTorch: An imperative style, high-performance deep learning library,'' in Adv. Neural Inf. Process. Syst. (NeurIPS), 2019, arXiv:1912.01703.
\bibitem{kalamkar2019study} D. Kalamkar \emph{et al.}, ``A study of BFLOAT16 for deep learning training,'' arXiv:1905.12322, 2019.
\bibitem{recht2019imagenet} B. Recht, R. Roelofs, L. Schmidt, and V. Shankar, ``Do ImageNet classifiers generalize to ImageNet?,'' arXiv:1902.10811, 2019.
\bibitem{beyer2020done} L. Beyer, O. J. H{\'e}naff, A. Kolesnikov, X. Zhai, and A. van den Oord, ``Are we done with ImageNet?,'' arXiv:2006.07159, 2020.
\bibitem{raji2021everything} I. D. Raji, E. M. Bender, A. Paullada, E. Denton, and A. Hanna, ``AI and the everything in the whole wide world benchmark,'' in Adv. Neural Inf. Process. Syst. Datasets and Benchmarks Track, 2021, arXiv:2111.15366.
\bibitem{dehghani2021benchmark} M. Dehghani \emph{et al.}, ``The benchmark lottery,'' arXiv:2107.07002, 2021.
\bibitem{henderson2018deep} P. Henderson, R. Islam, P. Bachman, J. Pineau, D. Precup, and D. Meger, ``Deep reinforcement learning that matters,'' in Proc. AAAI Conf. Artif. Intell., 2018, arXiv:1709.06560.
\bibitem{lipton2018troubling} Z. C. Lipton and J. Steinhardt, ``Troubling trends in machine learning scholarship,'' arXiv:1807.03341, 2018.
\bibitem{musgrave2020metric} K. Musgrave, S. Belongie, and S.-N. Lim, ``A metric learning reality check,'' arXiv:2003.08505, 2020.
\bibitem{guo2017calibration} C. Guo, G. Pleiss, Y. Sun, and K. Q. Weinberger, ``On calibration of modern neural networks,'' in Proc. Int. Conf. Mach. Learn. (ICML), 2017, arXiv:1706.04599.
\bibitem{koh2021wilds} P. W. Koh \emph{et al.}, ``WILDS: A benchmark of in-the-wild distribution shifts,'' arXiv:2012.07421, 2020.
\bibitem{bewley2016simple} A. Bewley, Z. Ge, L. Ott, F. Ramos, and B. Upcroft, ``Simple online and realtime tracking,'' in Proc. IEEE Int. Conf. Image Process. (ICIP), 2016, arXiv:1602.00763.
\bibitem{loshchilov2017decoupled} I. Loshchilov and F. Hutter, ``Decoupled weight decay regularization,'' in Proc. Int. Conf. Learn. Represent. (ICLR), 2019, arXiv:1711.05101.
\bibitem{lv2018unsupervised} J. Lv, W. Chen, Q. Li, and C. Yang, ``Unsupervised cross-dataset person re-identification by transfer learning of spatial-temporal patterns,'' in Proc. IEEE/CVF Conf. Comput. Vis. Pattern Recognit. (CVPR), 2018, arXiv:1803.07293.
\bibitem{ren2021learning} M. Ren, L. He, X. Liao, W. Liu, Y. Wang, and T. Tan, ``Learning instance-level spatial-temporal patterns for person re-identification,'' in Proc. IEEE/CVF Int. Conf. Comput. Vis. (ICCV), 2021, arXiv:2108.00171.
\bibitem{shu2021large} X. Shu \emph{et al.}, ``Large-scale spatio-temporal person re-identification: Algorithms and benchmark,'' \emph{IEEE Trans. Circuits Syst. Video Technol.}, 2021, arXiv:2105.15076.
\bibitem{cho2017distance} Y.-J. Cho and K.-J. Yoon, ``Distance-based camera network topology inference for person re-identification,'' arXiv:1712.00158, 2017.
\bibitem{kim2024object} H.-G. Kim, Y.-H. Moon, and Y.-J. Cho, ``Object re-identification via spatial-temporal fusion networks and causal identity matching,'' arXiv:2408.05558, 2024.
\bibitem{teng2022high} H. Teng, T. He, Y. Guo, and G. Ding, ``A high-accuracy unsupervised person re-identification method using auxiliary information mined from datasets,'' arXiv:2205.03124, 2022.
\bibitem{peng2016unsupervised} P. Peng \emph{et al.}, ``Unsupervised cross-dataset transfer learning for person re-identification,'' in Proc. IEEE/CVF Conf. Comput. Vis. Pattern Recognit. (CVPR), 2016, pp. 1306--1315, doi:10.1109/CVPR.2016.146.
\bibitem{farenzena2010person} M. Farenzena, L. Bazzani, A. Perina, V. Murino, and M. Cristani, ``Person re-identification by symmetry-driven accumulation of local features,'' in Proc. IEEE/CVF Conf. Comput. Vis. Pattern Recognit. (CVPR), 2010, pp. 2360--2367, doi:10.1109/CVPR.2010.5539926.
\bibitem{zheng2011person} W.-S. Zheng, S. Gong, and T. Xiang, ``Person re-identification by probabilistic relative distance comparison,'' in Proc. IEEE/CVF Conf. Comput. Vis. Pattern Recognit. (CVPR), 2011, pp. 649--656, doi:10.1109/CVPR.2011.5995598.
\bibitem{li2014deepreid} W. Li, R. Zhao, T. Xiao, and X. Wang, ``DeepReID: Deep filter pairing neural network for person re-identification,'' in Proc. IEEE/CVF Conf. Comput. Vis. Pattern Recognit. (CVPR), 2014, pp. 152--159, doi:10.1109/CVPR.2014.27.
\bibitem{koestinger2012large} M. Kostinger, M. Hirzer, P. Wohlhart, P. M. Roth, and H. Bischof, ``Large scale metric learning from equivalence constraints,'' in Proc. IEEE/CVF Conf. Comput. Vis. Pattern Recognit. (CVPR), 2012, pp. 2288--2295, doi:10.1109/CVPR.2012.6247939.
\bibitem{liao2015person} S. Liao, Y. Hu, X. Zhu, and S. Z. Li, ``Person re-identification by local maximal occurrence representation and metric learning,'' in Proc. IEEE/CVF Conf. Comput. Vis. Pattern Recognit. (CVPR), 2015, pp. 2197--2206, doi:10.1109/CVPR.2015.7298832.
\bibitem{gray2008viewpoint} D. Gray and H. Tao, ``Viewpoint invariant pedestrian recognition with an ensemble of localized features,'' in \emph{Lecture Notes in Computer Science}, 2008, pp. 262--275, doi:10.1007/978-3-540-88682-2\_21.
\bibitem{makris2004bridging} D. Makris, T. Ellis, and J. Black, ``Bridging the gaps between cameras,'' in Proc. IEEE Comput. Soc. Conf. Comput. Vis. Pattern Recognit. (CVPR), 2004, pp. 205--210, doi:10.1109/CVPR.2004.1315165.
\bibitem{javed2008modeling} O. Javed, K. Shafique, Z. Rasheed, and M. Shah, ``Modeling inter-camera space--time and appearance relationships for tracking across non-overlapping views,'' \emph{Comput. Vis. Image Underst.}, vol. 109, 2008, pp. 146--162, doi:10.1016/j.cviu.2007.01.003.

\end{thebibliography}
